\documentclass[twoside,11pt]{article}

\usepackage{jmlr2e}

\usepackage{hhline}
\usepackage{amsmath}
\usepackage[title]{appendix}
\usepackage{float}

\ewrlheading{14}{2018}{October 2018, Lille, France}{H\'el\`ene Plisnier, Denis Steckelmacher, Tim Brys, Diederik M. Roijers, Ann Now\'e}

\ShortHeadings{Directed Policy Gradient for Safe Reinforcement Learning with Human Advice}{Plisnier, Steckelmacher, Brys, Roijers and Now\'e} 
\firstpageno{1}

\begin{document}

\title{Directed Policy Gradient for Safe Reinforcement Learning with Human Advice}

\author{
	H\'el\`ene Plisnier,\textsuperscript{1}
	Denis Steckelmacher,\textsuperscript{1}
    Tim Brys,\textsuperscript{1}
	Diederik M. Roijers,\textsuperscript{2}
	Ann~Now\'e,\textsuperscript{1} \\
	\textsuperscript{1} Vrije Universiteit Brussel \\
	\textsuperscript{2} Vrije Universiteit Amsterdam
}

       

\maketitle

\begin{abstract}
Many currently deployed Reinforcement Learning agents work in an environment shared with humans, be them co-workers, users or clients. It is desirable that these agents adjust to people's preferences, learn faster thanks to their help, and act safely around them. We argue that most current approaches that learn from human feedback are unsafe: rewarding or punishing the agent a-posteriori cannot immediately prevent it from wrong-doing. In this paper, we extend Policy Gradient to make it robust to external directives, that would otherwise break the fundamentally on-policy nature of Policy Gradient. Our technique, Directed Policy Gradient (DPG), allows a teacher or backup policy to override the agent before it acts undesirably, while allowing the agent to leverage human advice or directives to learn faster. Our experiments demonstrate that DPG makes the agent learn much faster than reward-based approaches, while requiring an order of magnitude less advice.
\end{abstract}
.

\section{Introduction}
To promote harmonious human-agent cohabitation, it is crucial for Reinforcement Learning agents to conform and self-adjust to their users' preferences and requests quickly. Learning from human intervention not only allows users to customize the behavior of their agent, but can also help agents learn faster. Most current techniques learn from human-delivered feedback: shaping the environmental reward by mixing it with human reward \citep{Thomaz2006,Knox2010}, learning solely from human reward \citep{deeptamer2017,Christiano2017,Pilarski2017}, or directly shaping the agent's policy by mixing it with a human policy extracted from human feedback \citep{Griffith2013,Macglashan2017}. Feedback-based methods can effectively improve the performance of the agent; however, it does not allow the teacher to immediately deflect the agent's behavior in case it acts undesirably or unsafely.

In this paper, we extend Policy Gradient to allow human-provided advice to change which actions are selected by the agent. Our main contribution, Directed Policy Gradient (DPG), makes it possible for an advisory policy $\pi_H$ to directly influence the actions selected by a Policy Gradient agent, a practice originally restricted to off-policy algorithms only \citep{Fernandez2006}. Moreover, it empowers users by allowing them to take control of the agent whenever they judge necessary. In contrast to value-based methods such as SARSA and Q-Learning, Policy Gradient is able to ignore deterministic (i.e., orders) and stochastic (i.e., soft suggestions) incorrect advice, which is crucial in ensuring that the agent learns the best-possible policy regardless of the advice it receives (see our in-depth discussion in Section \ref{append:PGvsQvsSarsa}). We empirically demonstrate the properties of DPG on a challenging navigation task, and compare it to reward-shaping. Finally, although we consider here that the advisory fixed policy comes from a human, DPG is generalizable to any source of advisory policies, such as other agents, backup policies, or safe but sub-optimal expert policies.

\section{Background}
\label{sec:background}

In this section, we formally introduce Markov Decision Processes (MDPs), Policy Gradient, and how probability distributions (in our case, policies) are combined. 

\subsection{Markov Decision Processes}
\label{sec:background_mdp}

A discrete-time Markov Decision Process (MDP) \citep{Bellman1957} with discrete actions is defined by the tuple $\langle S, A, R, T \rangle$: a possibly-infinite set $S$ of states; a finite set $A$ of actions; a reward function $R(s_t, a_t, s_{t+1}) \in \mathcal{R}$ returning a scalar reward $r_t$ for each state transition; and a transition function $T(s_{t+1} | s_t, a_t) \in [0, 1]$ taking as input a state-action pair $(s_t, a_t)$ and returning a probability distribution over new states $s_{t+1}$.

A stochastic stationary policy $\pi$, with $\pi(a_t | s_t) \in [0, 1]$ the probability of action $a_t$ to be taken in state $s_t$, maps each state to a probability distribution over actions. At each time-step, the agent observes $s_t$, selects $a_t$ sampled from the state-dependent probability distribution over actions $\pi(s_t)$, then observes $r_{t+1}$ and $s_{t+1}$. The $(s_t, a_t, r_{t+1}, s_{t+1})$ tuple is called an \emph{experience} tuple. An optimal policy $\pi^*$ maximizes the expected cumulative discounted reward $E_{\pi^*}[\sum_t \gamma^t r_t]$. The goal of the agent is to find $\pi^*$ based on its experiences within the environment.

\subsection{Policy Gradient}
\label{sec:background_pg}

Policy Gradient methods \citep{Williams1992,Sutton2000} explicitly learn a policy $\pi_\theta$ parametrized by a weights vector $\theta$. The agent's goal is to maximize the expected cumulative discounted reward $E_{\pi}[\sum_t \gamma^t r_t]$, which consists in minimizing the following loss \citep{Sutton2000}:

\begin{align}
    \label{eq:pg}
    \mathcal{L}(\pi) &= -\sum\limits_{t=0}^{T} \mathcal{R}_t \log (\pi_{\theta} (a_t | s_t))
\end{align}

\noindent
where $a_t \sim \pi(s_t)$ is the action executed at time $t$. The return $\mathcal{R}_t = \sum_{\tau=t}^{T} \gamma^{\tau} r_{\tau}$, with $r_{\tau} = R(s_{\tau}, a_{\tau}, s_{\tau+1})$, is a simple discounted sum of future rewards. Intuitively, the loss leads to the probability of past actions with the highest return to be executed more often in the future, leading to a constantly improving policy. At every training epoch, experiences are used to compute the gradient $\frac{\partial L}{\partial \theta}$ of Equation \ref{eq:pg}, then the weights of the policy are adjusted one step in the direction of the gradient. Policy Gradient is said to be strongly \emph{on-policy}: it only converges if the actions executed by the agent are drawn from the exact probability distribution produced by $\pi$, which prevents any form of explicit exploration or advice.\footnote{Our testing revealed that even PPO (with discrete actions), much more robust and advanced than simple Policy Gradient, diverges when the slightest amount of off-policy directives override its behavior.} Our main contribution, DPG, overcomes this issue by allowing advice to directly influence $\pi(s_t)$, without any convergence issue. 

\subsection{Mixing Policies}
\label{sec:background_mixing}

Policy Shaping allows the agent's learned policy $\pi_L(s_t)$ to be influenced by another advisory fixed policy $\pi_H(s_t)$. The agent then samples actions from the mixture of the two policies, $\pi(s_t)$, as follows \citep{Griffith2013}:

\begin{align} 
    \label{eq:ps}
    a_t \sim \pi(s_t) = \frac{\pi_L(s_t) \, \times \, \pi_H(s_t)} {\sum_{a \in A} \pi_L(a|s_t) \, \times \, \pi_H(a|s_t)}
\end{align}

Even if first proposed for integrating collected human feedback in the learning process \citep{Griffith2013}, this simple method can be applied to a larger variety of problems. In addition to a human teacher, the advisory policy $\pi_H(s_t)$ can be a backup policy that prevents the agent from behaving undesirably, another's agent policy from which knowledge is transferred, or a policy to imitate. In this paper, we consider the setting where a (simulated) human provides deterministic advice to the agent, which forces its policy $\pi$ to choose certain actions.

\section{Directed Policy Gradient}
\label{sec:contribution}

DPG assumes a parametric policy represented by a neural network with two inputs: the state $s_t$ and an \emph{advice} $\pi_H$. The state is used to compute a probability distribution $\pi_L$ over actions, using dense, convolutional, or any other kind of trainable layers. Then, the output $\pi$ of the network is computed by element-wise multiplying $\pi_L$ and $\pi_H$ (see Section \ref{sec:background_mixing}), resulting in a mixture of the agent's policy and the advisory policy, from which actions can be directly sampled. In this neural architecture, inspired from how variable action-spaces are implemented in \cite{Steckelmacher2018}, the behavior of the agent can be directly altered by a teacher or a backup policy, while conventional Policy Gradient requires the actions taken to be exclusively sampled from the policy $\pi$ \citep{Sutton2000}. Moreover, our experimental results in Section \ref{sec:experiments} demonstrate that DPG is able to leverage advice to learn faster, instead of merely obeying instructions and then learning as if nothing happened.

More specifically, the policy over actions $\pi$ is represented as a feed-forward neural network (with one hidden layer of 100 neurons in our experiments), trained using Policy Gradient \citep{Sutton2000} and the Adam optimizer \citep{Kingma2014}. The neural network $\pi$ takes two inputs: $\mathbf{s}$ the state observed from the environment, and a $\mathbf{\pi_H}$ probability distribution over actions used for Policy Shaping. If no advice is available, $\mathbf{\pi_H}$ is set to the uniform distribution, which cancels its effect and allows Policy Gradient to learn as usual. The output of the $\pi$ network is computed as follows:

\begin{align*} 
    \mathbf{h_1} &= \tanh(\mathbf{W_1}\mathbf{s} + \mathbf{b_1}), \\
    \mathbf{\hat{y}} &= \sigma(\mathbf{W_2 h_1} + \mathbf{b_2}) \circ \mathbf{\pi_H}, \\
    \mathbf{y} &= \frac{\mathbf{\hat{y}}} {\mathbf{1^T \hat{y}}},
\end{align*}

with $\mathbf{W_i}$ and $\mathbf{b_i}$ the weights and biases of layer $i \in [1,2]$, $\sigma$ the sigmoid activation function, and $\circ$ denoting the element-wise product of two vectors. The network is trained using the standard Policy Gradient loss shown in Equation \ref{eq:pg} \citep{Sutton2000}. We now demonstrate the applicability of DPG to various settings, and illustrate how much less human interventions are needed for DPG to be useful, compared to reward-based approaches.

\section{Ignoring phony advice} 
\label{append:PGvsQvsSarsa}

\begin{figure}
	\centering
	\includegraphics{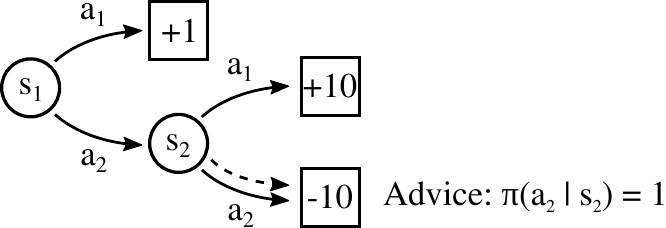}
	\caption{2-state and 2-action environment. Incorrect deterministic advice forces the agent, in state $s_2$, to take the sub-optimal action $a_2$. In this setting, Q-Learning still associates a large Q-Value to $s_2$, while Policy Gradient and SARSA are able to learn to execute $a_1$ in $s_1$, in order to avoid the consequences of bad advice.}
    \label{fig:phony}
\end{figure}

We illustrate, using a simple example in Figure \ref{fig:phony}, that Policy Gradient, in contrast to value-based methods such as Q-learning and SARSA, can ignore phony advice. This must be shown for two cases: either the erroneous advice is deterministic (i.e., the agent has no choice but to follow the advice), or stochastic (i.e., the advised action is non-mandatory).

\subsection{Deterministic Advice}

Deterministic advice is such that the advisory fixed policy $\pi_H(s_t)$ has a 1 for a particular action (the one that the agent is forced to take), and a 0 for all other actions. Consider the two-states and two-actions environment shown in Figure \ref{fig:phony}. In each of the two states $s_1, s_2$, the same two actions are available to the agent, $a_1$ and $a_2$. In state $s_1$, choosing action $a_1$ results in a positive reward of 1; choosing $a_2$ results in a zero reward, and makes the agent transition to state $s_2$. In $s_2$, choosing $a_1$ results in a positive reward of 10, while choosing $a_2$ results in a negative reward of $-10$. Without deterministic advice, Q-learning, SARSA and Policy Gradient learn the optimal policy, which is to select $a_2$ in $s_1$, then $a_1$ in $s_2$, hence collecting the +10 reward. Now, consider that phony deterministic advice is introduced, forcing the agent to select $a_2$ in $s_2$. Since the agent has no choice but to follow the deterministic advice once in $s_2$, the only way to avoid the -10 reward is to not transition to $s_2$, by choosing action $a_1$ in $s_1$.

Let's consider the Q-Learning and SARSA update rules:

\begin{align*} 
     Q^Q_{k+1}(s_t, a_t) &= Q^Q_k + \alpha(r_t + \max_{a'} Q^Q_k(s_{t+1}, a') - Q^Q_k(s_t, a_t)) \\
     Q^S_{k+1}(s_t, a_t) &= Q^S_k + \alpha(r_t + Q^S_k(s_{t+1}, a_{t+1}) - Q^S_k(s_t, a_t))
\end{align*}

If we apply these rules until convergence, and focus on state $s_1$, we obtain two Q-Values for Q-Learning, and two for SARSA:

\begin{align*} 
     Q^Q(s_1, a_1) &= 1 \\
     Q^Q(s_1, a_2) &= 0 + \max_{a'}Q^Q(s_2, a') &= 10 \\
     Q^S(s_1, a_1) &= 1 \\
     Q^S(s_1, a_2) &= 0 + Q^S(s_2, \overbrace{a_2}^{advice}) &= -10
\end{align*}

Hence, even though it is systematically forced to collect the negative reward of -10 in state $s_2$, the greedy policy of Q-Learning keeps on selecting action $a_2$ in state $s_1$. On the other hand, SARSA is able to learn the effect of the advice, which leads to the better $a_1$ action to be selected. Policy Gradient also learns the correct policy, because it relies on returns. In state $s_1$, $R = 1$ for action $a_1$, and $R = 0 - 10$ for action $a_2$. The learned policy, by maximizing the return, therefore converges towards deterministically choosing $a_1$ in $s_1$. 

\subsection{Stochastic Advice}

The previous sub-section shows that Q-Learning is not robust to deterministic incorrect advice. We now consider the stochastic advice setting, in which a non-zero probability is assigned to all the actions, and show that SARSA fails to robustly learn the optimal policy, while Policy Gradient manages to do so.

Value-based methods require explicit exploration, usually $\epsilon$-Greedy or Softmax. That explicit exploration, even for arbitrarily small epsilons or temperatures, prevents the policy followed by the agent from becoming fully deterministic. Without loss of generality, we now focus on SARSA with $\epsilon$-Greedy. Let's consider a one-state $s$ and two-actions $a_1, a_2$ environment; choosing action $a_1$ results in a reward of 1; choosing $a_2$ results in a reward of 0. We denote that as a return vector $\{1, 0\}$. An $\epsilon$-Greedy agent easily learns a policy $\pi_L(s) = \{1-\epsilon, \epsilon\}$, that is mixed with a phony stochastic advice $\pi_H(s) = \{0.01, 0.99\}$. Given the policy mixing Equation \ref{eq:ps} (in Section \ref{sec:background_mixing}), $\pi_H(s)$ and $\pi_L(s)$ are mixed as follows:

\begin{align*} 
    \pi(s) &= \left\{ \frac{(1-\epsilon) \times 0.01}{(1-\epsilon) \times 0.01 + \epsilon \times 0.99}, \frac{\epsilon \times 0.99}{(1-\epsilon) \times 0.01 + \epsilon \times 0.99} \right\} \\
    &= \left\{ \frac{0.01 - 0.01 \times \epsilon}{0.01 + 0.98 \times \epsilon}, \frac{0.99 \times \epsilon}{0.01 + 0.98 \times \epsilon} \right\}
\end{align*}

Given an $\epsilon > 0$, the suboptimal action $a_2$, leading to a 0 reward, will be taken with a non-zero probability. For instance, that action will still be taken with a probability of 10\% if $\epsilon = 0.001$, an unrealistic $\epsilon$ value for any real-world application. Hence, Q-learning and SARSA are unable to ignore phony stochastic advice.

The policy learned by Policy Gradient, on the other hand, is deterministic when converged, resulting in $\pi_L(s) = \{1, 0\}$. Hence, the mixing of $\pi_L(s)$ with the stochastic advisory policy $\pi_H(s) = \{0.01, 0.99\}$ leads to $\pi(s) = \{1, 0\}$. This way, Policy Gradient shows robustness in the face of phony advice, in both its deterministic and stochastic form.

\section{Experiments}
\label{sec:experiments}




\begin{figure}
	\centering
	\includegraphics{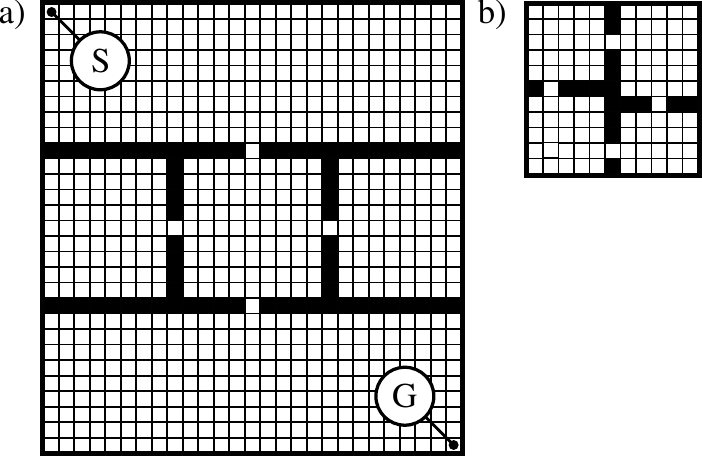}
	\caption{Comparison between a) our $29 \times 27$ Five rooms environment on which we evaluate our method and b) the $11 \times 11$ Four rooms environment \citep{options} which inspired Five Rooms. In a), black cells represent walls; the agent starts in cell S, and must reach the goal-cell G.}
	\label{fig:fiverooms}
\end{figure}

The evaluation of our method takes place in the Five Rooms grid world environment (see Figure \ref{fig:fiverooms}a), inspired from the well-known Four Rooms \citep{options} (in Figure \ref{fig:fiverooms}b). Its sheer size ($29 \times 27$ cells) and the fact that each room is accessible via tiny (one-cell wide) doors make exploration difficult. The agent starts in the top-left corner of the grid, and must reach the bottom right corner, where it receives a reward of $+100$. The agent's policy chooses amongst 5 macro-actions: one per door, driving the agent directly to that door from any cell in a neighboring room; the fifth macro-action goes to the goal from any cell of the bottom room. If the chosen macro-action is defined for the agent's current position (for example, considering one of the 4 door-macro-actions, if the agent is in a room directly accessible from that door), it leads directly to its destination. Otherwise, the macro-action moves the agent to a random neighboring cell before returning control to the policy over macro-actions. As a result, the agent visits potentially every cell of the grid, instead of just hopping from door to door, which keeps the problem challenging. After having executed a macro-action, the agent receives the sum of the rewards associated to each individual cell ($-0.1$ for every non-goal cell). The episode terminates either once the goal has been reached, or after 500 unfruitful time-steps. The optimal policy takes $54$ time-steps to reach the goal, and obtains a cumulative reward of $100 - 54 \times (-0.1) = 94.6$. Our use of a complicated environment with sparse rewards, and of macro-actions, is motivated by the nature of real-world robotic tasks in human-populated environments, that are big, complex, and already rely on macro-actions. 

In our experiments, we simulate human-provided advice using a deterministic function. To design this function, we got inspiration from a small sample of actual people giving advice to a learner. Before the agent chooses a macro-action, the human teacher will tell it which one to choose with an availability probability $0 \le L \le 1$. The probability of advice to be correct is modeled by $0 \le P(right) \le 1$; a \emph{wrong advice} consists in the teacher advising the agent to take the door leading to the middle left room, which takes the agent away from the direct path to the goal. Human advice is represented as a probability distribution over macro-actions, with a probability 1 for the action that the human wants the agent to execute, and 0 for every other action. The $\pi_H(s_t)$ advice is then mixed to the learned policy $\pi_L(s_t)$, as detailed in Section \ref{sec:contribution}.

\subsection{Directed Policy Gradient and Reward Shaping}
\label{subsec:expcompare}

\begin{figure}
	\centering
	\includegraphics[width=0.49\textwidth]{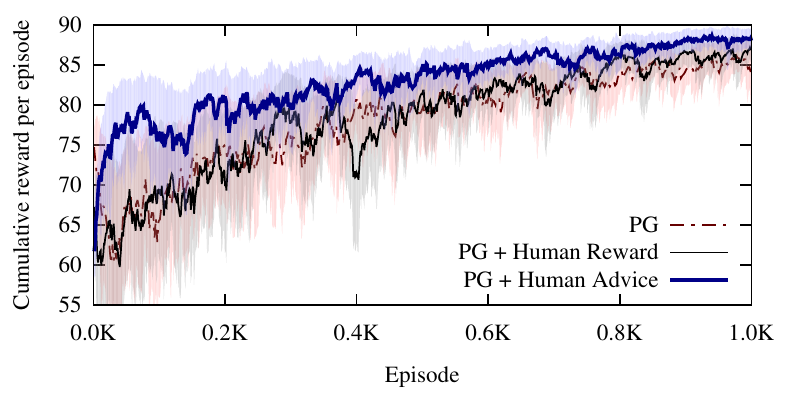}
    \includegraphics[width=0.49\textwidth]{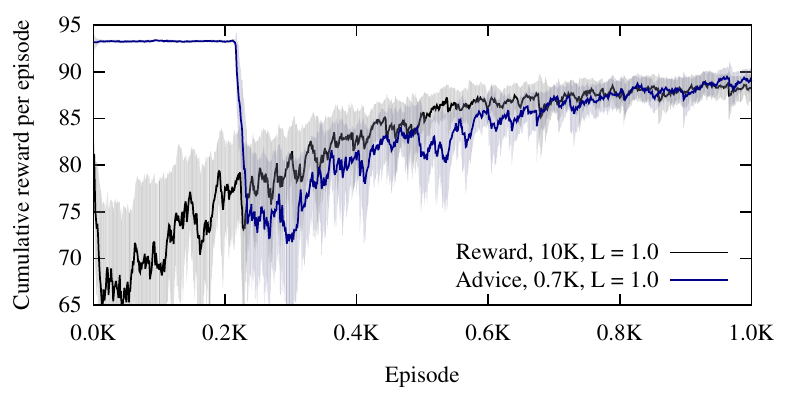}
	\caption{\textbf{Left:} comparison between Policy Gradient (PG) with human advice (PG + Human Advice) and with human reward (PG + Human Reward). Both human advice and reward are given to the agent with a probability $L=0.05$, corresponding to roughly 1000 human interventions for 1000 episodes. Human advice leads to the highest returns, and improves the performance of the agent right from the start. \textbf{Right:} 700 pieces of human advice, with $L = 1$, versus 10,000 reward-based punishments, with $L = 1$. Even though a far higher amount of reward is provided, human advice eventually matches, then slightly exceeds human reward.}
	\label{fig:PGvsHPvsHR}
\end{figure}

We compare DPG to simple Policy Gradient with reward shaping, and Policy Gradient without any form of human intervention. Human-provided reward with probability $L = 0.05$ (resulting in approximatively 1000 human interventions over 1000 episodes) is simulated using a deterministic function. Knowing the optimal policy over macro-actions, it rewards (by giving 0) or punishes (by giving -5) the agent if the macro-action chosen is the correct one according to the agent's current position. The numerical human reward is then added to the environmental reward. Although it is not potential-based, this does not prevent the agent from learning the task; and making it potential-based \citep{Harutyunyan2015} would require modeling the human's reward function, which might alter it. Human advice is given to DPG with the same probability $L = 0.05$, which also results in about 1000 human interventions over 1000 episodes.

Results in Figure \ref{fig:PGvsHPvsHR}, left, show that, in the beginning, the performance of the agent is not improved by human reward (p=0.419); however performance seems to be slightly improved towards the end of the 1000 episodes (p=0.007). On the other hand, DPG leads to much higher returns early on (p=1.369e-03), and continues to dominate reward shaping and simple Policy Gradient until the end of the 1000 episodes (p=2.396e-09). This experiment shows that, in contrast to reward-based methods, even a small amount of human advice significantly improves the agent's performance. The next experiments explore the behavior of DPG when various amounts of human advice is given.

\subsection{Non-Stationary Advice}
\label{subsec:expstop}

In many real-world settings, humans are not always available to help the agent learn. In Figure \ref{fig:PGvsHPvsHR}, right, we compare human advice, interrupted after 700 interventions, to human reward, interrupted after 10,000 punishments (so, much more than 10,000 time-steps during which the human had to watch the agent). This mimics an intense training period, followed by an abrupt interruption of human help, which assesses the robustness of the agent to sudden changes in human intervention. 

Once the 10,000 human punishments have been consumed, the performance of Policy Gradient with human reward slightly decreases, then plateaus until the end. Using only 700 pieces of advice, DPG manages to learn a policy marginally better than the one obtained with reward shaping, even if reward shaping requires one order of magnitude more human dedication. Furthermore, DPG is simpler to implement (see Section \ref{sec:contribution}) than reward shaping, which ideally requires a potential-based reward function to be designed \citep{Ng1999,Harutyunyan2015} to avoid biasing the learned policy.

\section{Conclusion and Future Work}

This paper presents Directed Policy Gradient, an extension of Policy Gradient that allows an advisory policy $\pi_H$ to directly influence the actions selected by the agent. We illustrate DPG in a human-agent cooperation setting, where the advisory policy is defined by a human. We show that DPG allows good policies to be learned from scarce advice, is robust to errors in the advice, and leads to higher returns than no advice, or reward-based approaches. Finally, although we used the example of a human advisory policy, and compared our work to other human-based approaches, it is important to note that any advisory policy can be used to shape a learner's policy, such as expert demonstrations, policies to be distilled from other agents, backup policies for Safe Reinforcement Learning, or a mix of all the above. DPG is therefore a straightforward, effective, and widely applicable approach to policy shaping.

\bibliography{bibli}

\end{document}